\documentclass{article}
\usepackage{spconf,amsmath,graphicx}

\graphicspath{{images/}}    % Specifies the directory where pictures are stored

\usepackage{dblfloatfix}    % for double column tables
\usepackage{setspace}       % for printing out drafts with "\doublespacing"
\usepackage{multirow}       % for multi-row tables
\usepackage{algorithm}
\usepackage{algpseudocode}

\usepackage[pagebackref=true,breaklinks=true,colorlinks,bookmarks=false]{hyperref}

\newcommand{\etal}{\mbox{\emph{et al.\ }}}
\newcommand{\etc}{\mbox{\emph{etc.\ }}}

\title{WordFence: Text Detection in Natural Images with Border Awareness}

\begin{document}
\name{Andrei Polzounov$^{1}$, Artsiom Ablavatski$^{2}$, Sergio Escalera$^{3}$, Shijian Lu$^{2}$, Jianfei Cai$^{4}$}
\vspace{-2cm}
\address{$^{1} $Universtitat Polit\`{e}cnica da Catalunya,  $^{2}$ A*STAR Institute for Infocomm Research,\\
$^{3}$Universitat de Barcelona and Computer Vision Center, $^{4}$Nanyang Technological University}

%\ninept
%
\maketitle
\vspace{-1cm}
\begin{abstract}
   In recent years, text recognition has achieved remarkable success in recognizing scanned document text. However, word recognition in natural images is still an open problem, which generally requires time consuming post-processing steps. We present a novel architecture for individual word detection in scene images based on semantic segmentation. Our contributions are twofold: the concept of WordFence, which detects border areas surrounding each individual word and a novel pixelwise weighted softmax loss function which penalizes background and emphasizes small text regions. WordFence ensures that each word is detected individually, and the new loss function provides a strong training signal to both text and word border localization. The proposed technique avoids intensive post-processing, producing an end-to-end word detection system. We achieve superior localization recall on common benchmark datasets - 92\% recall on ICDAR11 and ICDAR13 and 63\% recall on SVT. Furthermore, our end-to-end word recognition system achieves state-of-the-art 86\% F-Score on ICDAR13.
\end{abstract}
\begin{keywords}CNN, segmentation, word detection\end{keywords}

%%%%%%%%%%%%%%%%%%%%%%%%%%%%%%%%%%%%%%%%%%%%%%%%%%%%%%%%%%%%%%%%%%%%%%%%%%%%%%%%%%%%%%%%%%%%%%%%%%%%%%%%%%%%%%%%%%%%%%
% 1:  INTRODUCTION AND RELATED WORK
\section{Introduction and Related Work}\label{sec:Introduction}

Detection and recognition of text in natural images has long been an outstanding challenge in the computer vision and machine learning communities. Text recognition in the wild can provide context and semantic information for scene understanding, object classification or action recognition in images or video. The task has attracted the interest of many researchers \cite{gupta2016synthetic, jaderberg2016reading, tian2016detecting, tian2015text, zhong2016deeptext,he2016accurate}. Due to the difficulty of text detection in natural images, even state-of-the-art systems struggle with word localization because of the staggering variety of text sizes and fonts, potentially poor image quality, low contrast, image distortions, or presence of patterns visually similar to text such as: signs, icons or textures. Most works employ knowledge-based algorithms and heuristics in order to tackle these challenges. Common techniques include: text line extraction \cite{he2016accurate, tian2016detecting}, character candidate detection \cite{tian2015text} or secondary classifiers to remove false positive detections \cite{jaderberg2016reading}.

\begin{figure}[t]
\begin{center}
\includegraphics[width=1\linewidth]{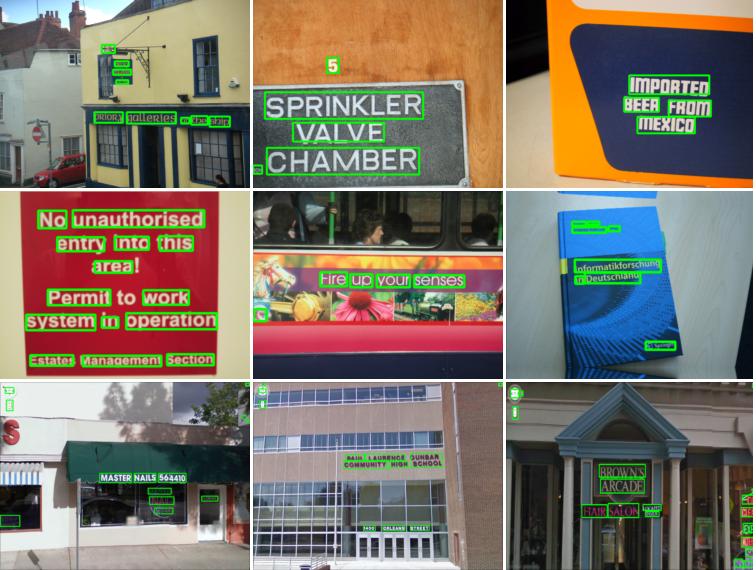}\\
\end{center}
   \caption{Word detection bounding box results on ICDAR2011 (top), ICDAR2013 (middle) and SVT (bottom) datasets. Bounding boxes are the output of the proposed method.}
\label{fig:SampleImages}
\end{figure}

Recent successes in computer vision are centered on convolutional neural networks (CNNs). Some of the problems being addressed with CNNs include: object-classification in natural images, pixelwise semantic segmentation \cite{long2015fully, chen2016deeplab, yu2015multi}, bounding box detection \cite{redmon2015you, liu2015ssd, sermanet2013overfeat} and text detection in scene images \cite{jaderberg2016reading, gupta2016synthetic, zhang2016multi, zhong2016deeptext, wang2012end, he2016text}.

A major limitation of CNNs is that networks have trouble taking different scales of images into account. Networks generally use max-pooling layers to reduce the search space for training - this operation reduces resolution and loses spatial information between different features. Yu and Koltun \cite{yu2015multi} argued that max-pooling does not maintain global scale information and propose dilated convolutions to increase the effective receptive field of convolutional operations. Other works tackled the scale problem with methods such as fully convolutional networks (FCNs) \cite{long2015fully} or with atrous convolutions \cite{sermanet2013overfeat, chen2016deeplab}. Another challenge addressed by CNNs is semantic segmentation - where each pixel in the image has to be matched to a specific label. Semantic segmentation has recently been enhanced by dilated convolutions \cite{yu2015multi}, FCNs \cite{long2015fully} and probabilistic graphical models \cite{chen2016deeplab}.

Traditionally text recognition has focused on documents and several optical character recognition (OCR) techniques have been developed for this task. Text recognition in scene imagery however, requires localizing the text first. Generally, text recognition works by first providing a "candidate bounding box" -- or a proposal for a single word or a word-line. This word proposal is then cropped out of the natural image and fed to a word recognition network which then matches words against an internal dictionary.

In the aforementioned scenario, text localization is considered to be the key task, since a well-cropped proposal can be fed to a word recognition system \cite{jaderberg2014synthetic}. Before CNNs, popular methods for text localization utilized computer vision techniques with hand-crafted feature descriptors. More recent works have used CNN features. However, all of these approaches have a limitation of feature driven engineering - there are simply too many edge cases to account for. The detectors generate a large amount of non-text false positives, requiring additional filtering techniques. Often, a number of post-processing steps is needed to reach a good performance.

With the prominence of deep learning, CNN based regression of candidate bounding boxes has started becoming utilized for filtering false positive candidates. Bounding box detection has been proposed in the context of object detection by works such as You Only Look Once (YOLO) \cite{redmon2015you}, Faster-RCNN (F-RCNN) \cite{ren2015faster} and SSD: Single Shot MultiBox Detector \cite{liu2015ssd}. Advances in semantic segmentation \cite{zhang2016multi,he2016accurate} have allowed dense prediction to provide input to bounding box regressors. Building on successful implementations of CNNs for semantic segmentation using FCNs for dense prediction \cite{long2015fully}, several researchers have introduced object localization via FCNs \cite{dai2016r}.

Early work by Zhang \etal \cite{zhang2016multi} used a semantic segmentation model to extract text proposals and refine them by applying hand-crafted heuristics. He \etal \cite{he2016accurate} improved on previous approaches by introducing a cascade of networks. Gupta \etal \cite{gupta2016synthetic} adapted YOLO's approach \cite{redmon2015you} for text detection and introduced SynthText - a new synthetic text dataset for training. Analogously, F-RCNN \cite{ren2015faster} was adapted for text recognition by Zhong \etal \cite{zhong2016deeptext} and Tian \etal \cite{tian2016detecting}. The former integrated the F-RCNN framework into a more powerful model. However, a large number of proposals needed to be filtered with a time consuming process. Tian \etal \cite{tian2016detecting} fused F-RCNN with a recurrent neural network (RNN), allowing the RNN to consider the proposals as a sequence.

Most current state-of-the-art region of interest (ROI) detectors like F-RCNN \cite{ren2015faster} use a variation of the following steps: propose bounding boxes, resample pixels of the ROI and then apply a second classifier to filter and improve proposals. In contrast with F-RCNN, our high quality segmentations allow us to extract accurate bounding box proposals directly from the segmentation. The segmentation maps are obtained by inference at different image scales, combining the results with an efficient voting mechanism. Merging the results from different scales helps to eliminate duplicate proposals for the same word and to remove most false positive detections.

Our proposed architecture is inspired by previously mentioned works, but it allows to perform bounding box detection in a single step. Instead of producing a highly non-linear bounding box coordinate prediction as in YOLO \cite{redmon2015you} and Faster-RCNN \cite{ren2015faster}, our network takes advantage of semantic segmentation to produce a dense pixel labeling map. Afterwards, word proposals are extracted from the given heat map in linear time (see Fig. \ref{fig:SampleImages} for examples).

% top of the 3rd page
\begin{figure*}[!t]
\begin{center}
\includegraphics[width=1\linewidth]{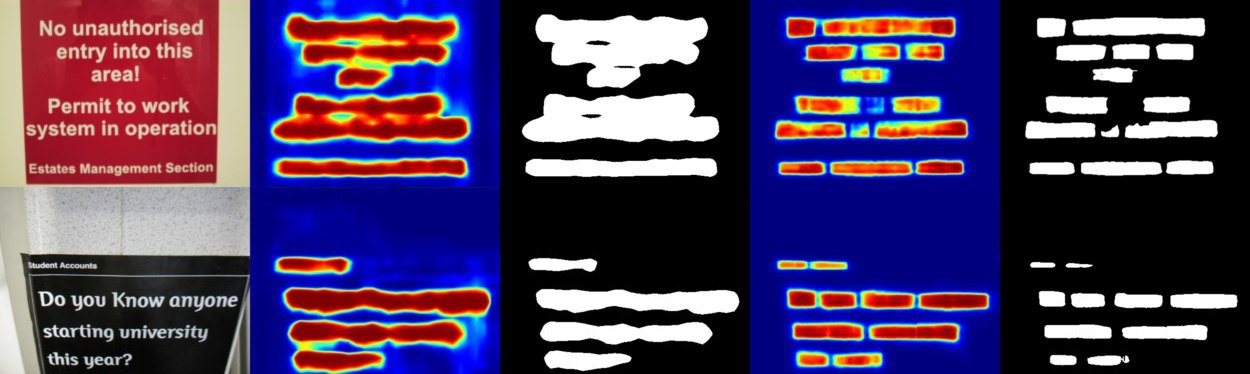}
\end{center}
    \caption{Segmentation comparisons with and without WordFence. First column from the left shows the original images. Second and third columns show the text position belief map and the resulting segmentation, respectively (trained with method outlined by Fisher and Koltun \cite{yu2015multi}). Last two columns show the belief map and the segmentation from our method. We found that an eight pixel border provides the best separation for most text sizes. Localizing words without WordFence causes individual words to bleed over into each other, which causes difficulty for posterior recognition.}
\label{fig:SegmentationsWithoutBorders}
\end{figure*}

%%%%%%%%%%%%%%%%%%%%%%%%%%%%%%%%%%%%%%%%%%%%%%%%%%%%%%%%%%%%%%%%%%%%%%%%%%%%%%%%%%%%%%%%%%%%%%%%%%%%%%%%%%%%%%%%%%%%%%
% 2: WordFence Detection Network
\section{WordFence Detection Network} \label{sec:Model}

Inspired by the success of deep CNNs with residual connections (ResNets), such as the one for semantic segmentation by Chen \etal \cite{chen2016deeplab}, our proposed WordFence Detection Network (WDN) takes advantage of recent deep learning research to produce highly accurate text detection results.
The network includes a ResNet-101 (introduced by He \etal \cite{he2015deep}), followed by a number of dilated convolutions \cite{yu2015multi} that add full image context to the final classification, before performing a bilinear interpolation on the resulting belief map. Afterwards, connected components are extracted. Each component represents a standalone word on the image which is further processed in the recognition step. Bounding boxes are then extracted from the connected components. Examples are shown in Fig. \ref{fig:SegmentationsWithoutBorders}.

\subsection{Word Localization as Semantic Segmentation}\label{sec:WordLocalizationAsSemanticSegmentation}
Object segmentation, has recently been considerably improved with the introduction of the deconvolutional layer \cite{long2015fully}, dilated convolutions (increasing effective receptive field) \cite{yu2015multi}, \etc Several published works \cite{zhang2016multi,he2016accurate} have adapted object segmentation for text localization. Segmentation for text localization, despite showing promising results, has had trouble distinguishing individual words from segmented images. Generally, post processing methods and heuristics were applied to refine word localization results, or the task was not addressed at all as in the case of textline approaches.

\subsection{ResNet of Exponential Receptive Fields}
Recently ResNets have achieved great success in different computer vision tasks \cite{he2015deep,chen2016deeplab}, even surpassing human performance. Their structure allows ResNets to train very deep neural networks without a vanishing gradient.

In contrast to the semantic segmentation model introduced by Chen \etal \cite{chen2016deeplab}, we do not use parallel replications of ResNet-101 on different scales as it makes the network computationally expensive to train. Instead, we use three parallel convolutional layers of the same kernel size, but different dilation parameters. This way we transform the convolutional features into parallel segmentation maps of different receptive fields. Separate dilated convolutions allow us to enlarge the effective receptive field of the CNN. This context information improves the network's understanding of text at different scales. Dilated convolutions do not increase the number of parameters, ensuring that the model remains easy to train. Finally, the obtained parallel segmentation maps are fused together by element wise summation, providing the final segmentation map, which are then used for word extraction.

%-------------------------------------------------------------------------
\subsection{Weighted Softmax Loss Function}\label{sec:WeightedSoftmaxLossFunction}

A common loss function for training semantic segmentation networks is a pixelwise classification softmax loss. Such a function is appropriate for dense pixelwise labelling if there are many classes. For text localization, the pixelwise softmax loss tends to force the network to produce merged segmenations on the borders of words results such as the ones illustrated in Fig. \ref{fig:SegmentationsWithoutBorders}. Post processing techniques are required to enhance the segmentation bounding boxes in order to use them for text recognition. In order to overcome this problem, a simple and efficient technique is introduced: instead of a binary text/non-text classification we define the notion of a border for each separate word as a third class. The border acts as a penalization for training. The model is driven to surround each separate word with an artificial barrier, which greatly reduces the ease and computational cost of reading separate words. During inference, individual words are cleanly segmented from each other and can then be extracted using connected components analysis.

\begin{table*}[t]
\begin{minipage}{\textwidth}
\begin{center}
    \begin{tabular}{|l|c|c|c|c|c|c|c|c|c|}
    \hline
    \multicolumn{1}{|c|}{\multirow{3}{*}{\textbf{Model}}}  & \multicolumn{9}{c|}{\textbf{PASCAL VOC IoU = 0.5}}                                                                                                                                                                         \\ \cline{2-10}
    \multicolumn{1}{|c|}{}                       & \multicolumn{3}{c|}{\textbf{ICDAR11}}         & \multicolumn{3}{c|}{\textbf{ICDAR13}}         & \multicolumn{3}{c|}{\textbf{SVT}} \\ \cline{2-10}
    \multicolumn{1}{|c|}{}                       & Prec.         & Rec.          & F-score       & Prec.         & Rec.          & F-score       & Prec.         & Rec.          & F-score       \\ \hline
    Tian \etal \cite{tian2016detecting}          & 0.89          & 0.79          & 0.84          & 0.93          & 0.83          & \textbf{0.88} & -             & -             & -             \\
    Gupta \etal \cite{gupta2016synthetic}        & 0.78          & 0.63          & 70.0          & 0.78          & 0.63          & 0.70          & 0.47          & 0.45          & 0.46          \\
    Jaderberg \etal \cite{jaderberg2016reading}* & 0.89          & 0.68          & 77.4          & 0.89          & 0.68          & 0.77          & 0.59          & 0.49          & 0.54          \\
    Gupta \etal \cite{gupta2016synthetic}*       & \textbf{0.94} & 0.77          & \textbf{0.85} & \textbf{0.94} & 0.76          & 0.84          & \textbf{0.65} & 0.60          & \textbf{0.62} \\ \hline
    \textbf{WDN Recognition (ours)}              & 0.64          & \textbf{0.92} & 0.75          & 0.65          & \textbf{0.92} & 0.76          & 0.47          & \textbf{0.63} & 0.54          \\ \hline
    \end{tabular}
\end{center}
\caption{State-of-the-art comparison for word detection. Precision, Recall and F-Score are reported. Recall maximization was necessary for obtaining good word detection results. Methods marked with * use a multistage false-positive filtering process to increase precision, the code was not published thus the results are not directly comparable with ours.}
\label{table:TextLocalizationResults}
\end{minipage}
\end{table*}

\begin{algorithm}
    \label{alg:WeightedSoftmax}
    \begin{algorithmic}[1]
    \Require{Predicates after fusion \textbf{Pr}, ground truth labels \textbf{L}}
    \State  $probs \gets \textit{Softmax}(\textbf{Pr})$ \Comment{pixel probabilities}
    \State $m \gets \textit{NumberOfUniqueLabels}(\textbf{L})$
    \State $n_1,n_2,\dots,n_m \gets  \textit{CountsOfUniqueLabels}(\textbf{L})$ \Comment{get counts of each label on a ground truth image}
    \State $loss \gets -\sum \frac{1}{n_{gt}} \log(probs_{gt})$ \Comment{weighted loss calculation}
    \State $\textit{Backpropagate}(loss,\frac{1}{n_1},\frac{1}{n_2},\dots,\frac{1}{n_m})$ \Comment{loss backprop with normalization factors}
    \caption{Pixelwise Weighted Softmax Loss}
    \end{algorithmic}
\end{algorithm}

The number of text pixels in a text recognition dataset may not be balanced among labels and the vast majority of all pixels are simply background - networks tend to predict background everywhere. To solve this issue, we introduce a weighted normalization. The novel loss function automatically penalizes predictions for pixels which form the majority of a given image and emphasizes pixels which are fewer in number. This makes the loss function well constrained for the task of text segmentation. Weight normalization is applied in two places: loss calculation and loss backpropagation. Normalization factors are calculated on the fly and are inversely proportional to pixel counts of each class. The algorithm of the weighted softmax loss function is shown in Alg. \ref{alg:WeightedSoftmax}.

%%%%%%%%%%%%%%%%%%%%%%%%%%%%%%%%%%%%%%%%%%%%%%%%%%%%%%%%%%%%%%%%%%%%%%%%%%%%%%%%%%%%%%%%%%%%%%%%%%%%%%%%%%%%%%%%%%%%%%
% 3: EXPERIMENTS AND RESULTS
\section{Experiments and Results}\label{sec:Experiments}

\subsection{Datasets}\label{sec:Datasets}

Our model is trained and evaluated on a number of different text detection datasets. The COCO-Text dataset \cite{veit2016cocotext} is based on the earlier MS-COCO dataset for object classification \cite{lin2014microsoft}. SynthText \cite{gupta2016synthetic} consists of natural images with synthetic text labels. ICDAR 2011 \cite{shahab2011icdar} and ICDAR 2013 \cite{gomez2013multi} are common benchmark datasets from the International Conference of Document Analysis and Recognition. Street View Text dataset (SVT) \cite{wang2010word} was harvested from Google Street View images. We train our model on MS-COCO, finetune on SynthText and evaluate on ICDAR11, ICDAR13 and SVT for comparison with state-of-the-art methods.

\subsection{Word Localization Experiments}\label{sec:WordLocalizationExperiments}

For the evaluation of our word detection results we use a PASCAL VOC style protocol where a proposal with intersection-over-union (IoU) $\geq$ 0.5 is considered a positive detection. PASCAL VOC is suitable for detecting individual words as it penalizes areas covering multiple words.

Running the image inference at different scales produces different segmentation maps that need to be processed afterwards. When merging segmentations from different scales, the results will contain many duplicates and false positives, but recall will be high since true positives will likely have been found. We adopt a mechanism for merging segmentation maps of different scales before extracting the bounding boxes, while maintaining a high recall. We use a voting scheme to produce a final segmentation map. We upscale all segmentation maps and find labels that correspond to maximal class probabilities in the segmentation maps. We extract the probability values for the found labels and sum them up on corresponding channels producing the map of summed maximum probabilities from different scales. The final segmentation is obtained by finding labels with maximum probabilities on the combined map giving fewer false positives.

Table \ref{table:TextLocalizationResults} shows the performance of WDN on benchmark datasets. On average we improved recall by $15\%$ over the previous multi-scale detection method by Gupta \etal \cite{gupta2016synthetic}.

\subsection{End-to-end Word Detection and Recognition} \label{sec:EndToEndWordDetection}

Using ideal, single-word proposals recognition accuracy can be as high as $98\%$ \cite{jaderberg2016reading}. In order to show the effectiveness and quality of proposals we integrate our model with a state-of-the-art recognition model by Shi \etal \cite{shi2015end}. The recognition model consists of an RNN to recognize words of different length. Our word proposals are cropped out and evaluated with the recognition network. We followed the evaluation protocol outlined by Wang \etal \cite{wang2011end}, where all word proposals that are three characters long or less or those that contain non-alphanumeric characters are ignored. An IoU overlap of 0.5 is required for a positive detection. Results for common recognition dataset are illustrated in Table \ref{table:EndToEndRecognition}. Our detection network achieves state-of-the-art recall rates - ensuring good candidate words. This combined with the recognition module obtains very accurate results for end-to-end word recognition. The network outperforms results by Jaderberg \etal \cite{jaderberg2016reading} and is on par or better than Gupta \etal \cite{gupta2016synthetic}.

\begin{table}[t]
    \begin{center}
    \begin{tabular}{|l|c|c|c|}
    \hline
    \textbf{Model}                              & \textbf{Year} & \textbf{ICDAR11} & \textbf{ICDAR13} \\ \hline
    Neumann \etal \cite{neumann2012real}        & 2013          & 0.45             & -                \\
    Jaderberg \etal \cite{jaderberg2016reading} & 2015          & 0.69             & 0.76             \\
    Gupta \etal \cite{gupta2016synthetic}       & 2015          & \textbf{0.84}    & 0.85             \\ \hline
    \textbf{WDN Recognition}                    & 2016          & \textbf{0.84}    & \textbf{0.86}    \\ \hline
    \end{tabular}
    \end{center}
    \caption{Evaluation of end-to-end word recognition on ICDAR 2011 and 2013 datasets. Our work is compared against other methods. F-score is reported.}
    \label{table:EndToEndRecognition}
\end{table}

%%%%%%%%%%%%%%%%%%%%%%%%%%%%%%%%%%%%%%%%%%%%%%%%%%%%%%%%%%%%%%%%%%%%%%%%%%%%%%%%%%%%%%%%%%%%%%%%%%%%%%%%%%%%%%%%%%%%%%
% 4. CONCLUSION
\section{Conclusion}\label{sec:Conclusion}

We have presented the WordFence Detection Network. WDN relies on space between words to accurately split words using purely visual information for a wide variety of fonts, text sizes, scales, orientations and text languages. After segmenting an image proposal bounding boxes are extracted at multiple scales with high detection recall. Lastly, end-to-end word recognition achieves state-of-the-art results with 84 \% and 86 \% F-Score on ICDAR11 and ICDAR13, respectively. We obtain such high end-to-end scores by leveraging high quality proposals and high recall in the detection stage. Experimental results show that our approach achieves competitive performance on ICDAR11 and ICDAR13 without utilizing any heuristics or knowledge based approaches.\footnote{Work was partially supported by the Spanish project TIN2016-74946-P (MINECO/FEDER, UE) and CERCA Programme, Generalitat de Catalunya.}

%%%%%%%%%%%%%%%%%%%%%%%%%%%%%%%%%%%%%%%%%%%%%%%%%%%%%%%%%%%%%%%%%%%%%%%%%%%%%%%%%%%%%%%%%%%%%%%%%%%%%%%%%%%%%%%%%%%%%%
% 5. REFERENCES
\newpage
{\small
\bibliographystyle{IEEEbib}
\bibliography{refs.bbl}
}

\end{document}